\documentclass[runningheads]{llncs}
\usepackage{graphicx}
\usepackage{enumitem}
\usepackage{amsmath}
\usepackage{hyperref}
\usepackage{tikz}
\usepackage{pgf}
\usepackage{url}
\usepackage{amssymb} 
\usepackage{mathtools}

\usepackage{bibentry}   
\usepackage{doi}
\usepackage{appendix}
\usepackage{graphicx}
\usepackage[normalem]{ulem}
\useunder{\uline}{\ul}{}
\usepackage{multicol}
\usepackage{multirow}
\usepackage[linesnumbered,ruled,vlined]{algorithm2e}
\usepackage{float}
\newcommand{\equalcontribution}{\textsuperscript{*}}
\makeatletter
\renewcommand\thanks[1]{\renewcommand\thefootnote{*}\footnote{#1}\renewcommand\thefootnote{\@arabic\c@footnote}}
\makeatother
\begin{document}
\title{Wave-Mask/Mix: Exploring Wavelet-Based Augmentations for Time Series Forecasting}
\titlerunning{Wavelet-Based Augmentations for Time Series Forecasting}
% If the paper title is too long for the running head, you can set
% an abbreviated paper title here
%

\author{Dona Arabi\inst{\equalcontribution} \and
Jafar Bakhshaliyev\inst{\equalcontribution} \and
Ayse Coskuner\inst{\equalcontribution} \and Kiran Madhusudhanan\inst{\equalcontribution} \and Kami Serdar Uckardes\thanks{Equal Contribution}}

%

% First names are abbreviated in the running head.
% If there are more than two authors, 'et al.' is used.
\authorrunning{D. Arabi et al.}

\institute{University of Hildesheim, 31141 Hildesheim, Germany\\
\email{\{arabi, bakhshaliyev, coskuner, ueckardes\}@uni-hildesheim.de} \\ \email{\{kiranmadhusud\}@ismll.de}}
\maketitle              % typeset the header of the contribution

\begin{abstract}
Data augmentation is important for improving machine learning model performance when faced with limited real-world data. In time series forecasting (TSF), where accurate predictions are crucial in fields like finance, healthcare, and manufacturing, traditional augmentation methods for classification tasks are insufficient to maintain temporal coherence. This research introduces two augmentation approaches using the discrete wavelet transform (DWT) to adjust frequency elements while preserving temporal dependencies in time series data. Our methods, Wavelet Masking (WaveMask) and Wavelet Mixing (WaveMix), are evaluated against established baselines across various forecasting horizons. To the best of our knowledge, this is the first study to conduct extensive experiments on multivariate time series using Discrete Wavelet Transform as an augmentation technique~\cite{survey}. Experimental results demonstrate that our techniques achieve competitive results with previous methods. We also explore cold-start forecasting using downsampled training datasets, comparing outcomes to baseline methods. Code is available at \href{https://github.com/jafarbakhshaliyev/Wave-Augs}{https://github.com/jafarbakhshaliyev/Wave-Augs}.
\end{abstract}
\section{Introduction}

Synthetic data generation is crucial when real-world data is scarce or insufficient. Data augmentation creates additional training samples by applying different transformations to existing data. This expands the dataset to include a wider variety of data patterns, strengthening model resilience and improving its ability to generalize to new data~\cite{data_aug}. 

Time series forecasting is crucial in diverse fields such as finance, healthcare, meteorology, and manufacturing, and data augmentation techniques have become essential for tasks such as classification, forecasting, anomaly detection, and clustering~\cite{fr_aug}. Data augmentation (DA) is increasingly essential in the realm of time series forecasting due to its ability to enhance model performance and improve generalization. Time series data, unlike image data, presents unique characteristics in both the time and frequency domains, making its analysis more intricate. While current augmentation methods are tailored for classification purposes, it is crucial in the context of time series forecasting to emphasize diversity and consistency with the original temporal patterns. In Time Series Forecasting (TSF), data points within the look-back window and forecasting horizon serve as both the data and the label, enabling the modeling of complex temporal relationships by accurately capturing data within the specified time frame~\cite{fr_aug,staug}.

The existing fundamental augmentation techniques encompass scaling, flipping~\cite{basic}, window cropping or slicing~\cite{windowcrop}, Gaussian noise injection~\cite{gaussian}, dynamic time warping~\cite{survey}, and other methods. However, these methods have a significant limitation as they may introduce missing values or alter periodic patterns within the series, rendering them unsuitable for TSF research. Consequently, the precision required for augmented data-label pairs in TSF exceeds that needed for other time series analysis tasks~\cite{fr_aug}. STAug, a decomposition-based augmentation approach~\cite{staug}, is effective for TSF but is inefficient and memory-intensive, limiting its application to large datasets. One of the baseline methods is the frequency-domain augmentation technique, FrAug~\cite{fr_aug}, which motivated our study. Due to its utilization of Fourier transform to decompose the signal, a key drawback is its exclusive focus on frequency components at the expense of time resolution~\cite{fr_aug}. Our methods, employing wavelet decomposition, exhibit potential for superior results compared to alternative time-frequency analysis techniques. These methods, along with others for comparison, are depicted in Fig.~\ref{fig01}. To grasp the limitations of each approach, we consider a signal with unknown frequency components. The Fourier transform provides detailed information about frequency components over time, showcasing high resolution in the frequency domain while overlooking variations in the time domain. Short-time Fourier Transform (STFT) addresses this limitation by analyzing localized frequency variations in the time domain. Nonetheless, STFT's fixed resolution poses a challenge, as its window boundaries remain constant~\cite{wavmotiv,dwt}. Overcoming this limitation, the wavelet transform offers variable window sizes, providing high frequency domain resolution and low time domain resolution for small frequency values and large frequency values respectively~\cite{dwtexp,wavmotiv,dwtexp2}. Given this information, adjusting the signal's frequency components at different resolutions is expected to yield favorable outcomes.

\tikzset{every picture/.style={line width=0.75pt}} %set default line width to 0.75pt  

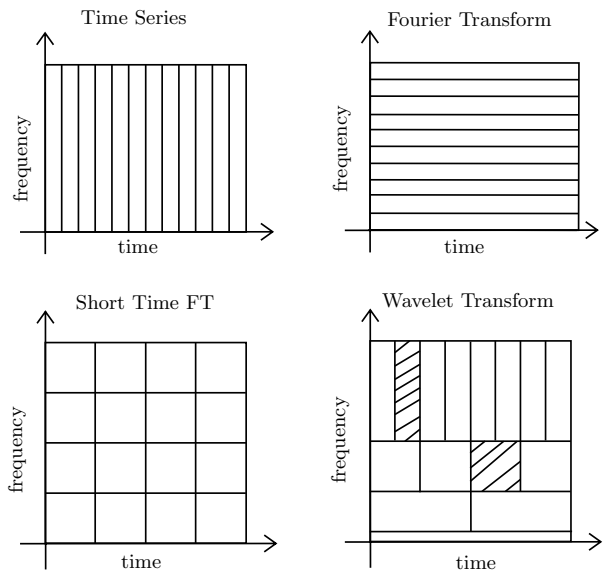
\begin{figure}[h!]
\centering
\resizebox{0.65\textwidth}{!}{%
\begin{tikzpicture}[x=0.75pt,y=0.75pt,yscale=-1,xscale=1]

%uncomment if require: \path (0,371); %set diagram left start at 0, and has height of 371

%Shape: Axis 2D [id:dp22694993501304084] 
\draw  (158.1,152.92) -- (308.99,152.79)(173.09,33.99) -- (173.2,166.12) (301.98,147.79) -- (308.99,152.79) -- (301.99,157.79) (168.09,40.99) -- (173.09,33.99) -- (178.09,40.98)  ;
%Shape: Rectangle [id:dp7994620226746767] 
\draw   (173.19,52.96) -- (293.17,52.96) -- (293.17,152.91) -- (173.19,152.91) -- cycle ;
%Straight Lines [id:da16643642622327648] 
\draw    (183,53) -- (183,153) ;
%Straight Lines [id:da5973026028749597] 
\draw    (193,53) -- (193,153) ;
%Straight Lines [id:da7788806936331576] 
\draw    (203,53) -- (203,153) ;
%Straight Lines [id:da005249434199233072] 
\draw    (213,53) -- (213,153) ;
%Straight Lines [id:da4958003614024926] 
\draw    (223,53) -- (223,153) ;
%Straight Lines [id:da41779925059501655] 
\draw    (233.18,52.93) -- (233.18,152.93) ;
%Straight Lines [id:da27569110850113177] 
\draw    (243,53) -- (243,153) ;
%Straight Lines [id:da3382001485161068] 
\draw    (253,53) -- (253,153) ;
%Straight Lines [id:da19227468816456073] 
\draw    (263,53) -- (263,153) ;
%Straight Lines [id:da25284651176519457] 
\draw    (273,53) -- (273,153) ;
%Straight Lines [id:da6474882462414455] 
\draw    (283,53) -- (283,153) ;
%Shape: Axis 2D [id:dp8501237676491376] 
\draw  (158.12,339.22) -- (310.99,339.09)(173.29,200.49) -- (173.42,354.62) (303.98,334.09) -- (310.99,339.09) -- (303.99,344.09) (168.29,207.49) -- (173.29,200.49) -- (178.29,207.48)  ;
%Shape: Rectangle [id:dp6120837724832282] 
\draw   (173.19,219) -- (293.17,219) -- (293.17,338.91) -- (173.19,338.91) -- cycle ;
%Straight Lines [id:da4233033028546074] 
\draw    (203,219) -- (203,339) ;
%Straight Lines [id:da6401647737551357] 
\draw    (173,309) -- (293,309) ;
%Straight Lines [id:da013244741275045602] 
\draw    (173,249) -- (293,249) ;
%Straight Lines [id:da39526907755234864] 
\draw    (173.18,278.95) -- (293.18,278.95) ;
%Straight Lines [id:da4974684043192117] 
\draw    (233.18,218.95) -- (233.18,338.95) ;
%Straight Lines [id:da2931939700859021] 
\draw    (263,219) -- (263,339) ;
%Shape: Axis 2D [id:dp8471911205381055] 
\draw  (352.1,151.92) -- (502.99,151.79)(367.09,32.99) -- (367.2,165.12) (495.98,146.79) -- (502.99,151.79) -- (495.99,156.79) (362.09,39.99) -- (367.09,32.99) -- (372.09,39.98)  ;
%Shape: Rectangle [id:dp8740905275725273] 
\draw   (491.94,51.87) -- (491.73,151.87) -- (367.14,151.7) -- (367.35,51.7) -- cycle ;
%Straight Lines [id:da12026644960163257] 
\draw    (491.93,61.87) -- (367.28,61.69) ;
%Straight Lines [id:da5424038524928938] 
\draw    (491.9,71.87) -- (367.26,71.69) ;
%Straight Lines [id:da4790779271709291] 
\draw    (491.88,82.87) -- (367.24,82.69) ;
%Straight Lines [id:da7576034299783532] 
\draw    (491.86,91.87) -- (367.21,91.69) ;
%Straight Lines [id:da9337500726632479] 
\draw    (491.84,101.87) -- (367.19,101.69) ;
%Straight Lines [id:da379581172721432] 
\draw    (491.9,112.05) -- (367.26,111.88) ;
%Straight Lines [id:da8216041393108162] 
\draw    (491.8,121.87) -- (367.15,121.69) ;
%Straight Lines [id:da8713984994305697] 
\draw    (491.77,130.87) -- (367.13,130.69) ;
%Straight Lines [id:da02522457167337855] 
\draw    (491.75,141.87) -- (367.11,141.69) ;
%Straight Lines [id:da6547172660885288] 
\draw    (490.73,151.87) -- (366.08,151.69) ;
%Shape: Axis 2D [id:dp5124402831831572] 
\draw  (352.12,338.22) -- (504.99,338.09)(367.29,199.49) -- (367.42,353.62) (497.98,333.09) -- (504.99,338.09) -- (497.99,343.09) (362.29,206.49) -- (367.29,199.49) -- (372.29,206.48)  ;
%Shape: Rectangle [id:dp7123779203820673] 
\draw   (367.19,218) -- (487.17,218) -- (487.17,337.91) -- (367.19,337.91) -- cycle ;
%Straight Lines [id:da21845851755837242] 
\draw    (367,308) -- (487,308) ;
%Straight Lines [id:da6267073982016074] 
\draw    (367.18,277.95) -- (487.18,277.95) ;
%Straight Lines [id:da5128312734886578] 
\draw    (382,218) -- (382,277.5) ;
%Straight Lines [id:da38328977024371413] 
\draw    (397,218) -- (397,308.5) ;
%Straight Lines [id:da20306903268402499] 
\draw    (412,218) -- (412,277.5) ;
%Straight Lines [id:da0916772438816793] 
\draw    (427.18,218.45) -- (427.42,331.93) ;
%Straight Lines [id:da4226475693440923] 
\draw    (442,218) -- (442,277.5) ;
%Straight Lines [id:da6135169633408759] 
\draw    (457,218) -- (457,308) ;
%Straight Lines [id:da7744837236831059] 
\draw    (487.84,331.87) -- (367,332) ;
%Straight Lines [id:da922823713982079] 
\draw    (472,218) -- (472,277.5) ;
%Straight Lines [id:da4211582446863993] 
\draw    (435.93,277.87) -- (427,285) ;
%Straight Lines [id:da15779511991706818] 
\draw    (456.93,300.87) -- (448,308) ;
%Straight Lines [id:da985081862993761] 
\draw    (447,278) -- (427,294) ;
%Straight Lines [id:da23985961840933356] 
\draw    (455,278) -- (427,302) ;
%Straight Lines [id:da6590700375221359] 
\draw    (457,290) -- (434,308) ;
%Straight Lines [id:da24823866186886767] 
\draw    (390.93,217.87) -- (382,225) ;
%Straight Lines [id:da6561198052042927] 
\draw    (397,224) -- (382,233) ;
%Straight Lines [id:da03689605577397814] 
\draw    (397,232) -- (382,241) ;
%Straight Lines [id:da8813324661012236] 
\draw    (397,238.75) -- (382,247.75) ;
%Straight Lines [id:da7974826328518321] 
\draw    (397,246) -- (382,255) ;
%Straight Lines [id:da021650449805376448] 
\draw    (397,253) -- (382,262) ;
%Straight Lines [id:da5495455588678391] 
\draw    (397,263.25) -- (382,272.25) ;
%Straight Lines [id:da6979239601729541] 
\draw    (396.93,270.87) -- (388,278) ;

% Text Node
\draw (215,156) node [anchor=north west][inner sep=0.75pt]   [align=left] {time};
% Text Node
\draw (152.71,136.55) node [anchor=north west][inner sep=0.75pt]  [rotate=-269.65] [align=left] {frequency};
% Text Node
\draw (410,156) node [anchor=north west][inner sep=0.75pt]   [align=left] {time};
% Text Node
\draw (343.71,130.55) node [anchor=north west][inner sep=0.75pt]  [rotate=-269.65] [align=left] {frequency};
% Text Node
\draw (149.71,310.55) node [anchor=north west][inner sep=0.75pt]  [rotate=-269.65] [align=left] {frequency};
% Text Node
\draw (217,344) node [anchor=north west][inner sep=0.75pt]   [align=left] {time};
% Text Node
\draw (193,19) node [anchor=north west][inner sep=0.75pt]   [align=left] {Time Series};
% Text Node
\draw (376,20) node [anchor=north west][inner sep=0.75pt]   [align=left] {Fourier Transform};
% Text Node
\draw (190,189) node [anchor=north west][inner sep=0.75pt]   [align=left] {Short Time FT};
% Text Node
\draw (417,345) node [anchor=north west][inner sep=0.75pt]   [align=left] {time};
% Text Node
\draw (342.71,304.55) node [anchor=north west][inner sep=0.75pt]  [rotate=-269.65] [align=left] {frequency};
% Text Node
\draw (373,188) node [anchor=north west][inner sep=0.75pt]   [align=left] {Wavelet Transform};

\end{tikzpicture}}
\caption{Illustration of time-frequency analysis methods~\cite{gabry2023}} \label{fig01}
\end{figure}

In this study, we introduce two innovative yet direct and simple data augmentation methods, termed as \textit{wavelet masking} (WaveMask) and \textit{wavelet mixing} (WaveMix). These techniques utilize the discrete wavelet transform (DWT) to obtain wavelet coefficients (both approximation and detail coefficients) by breaking down the signal and adjusting these coefficients, in line with modifying frequency components across different time scales. The DWT accomplishes this decomposition by employing a sequence of filtering and downsampling procedures, resulting in a structured representation of the signal at different levels of resolution. More extensive insights into DWT can be found in Appendix A. WaveMask selectively eliminates specific wavelet coefficients at each decomposition level, thereby introducing variability in the augmented data. Conversely, WaveMix exchanges wavelet coefficients from two distinct instances of the dataset, thereby enhancing the diversity of the augmented data.

The key contributions of this study include:
\begin{itemize}[leftmargin=1cm, label=$\bullet$]
    \item WaveMask and WaveMix techniques provide a richer understanding of a signal compared to FrAug methods, making them more competitive and straightforward.
    \item Our methods demonstrate superior performance in 12 out of 16 forecasting horizon tasks and rank second best in the remaining four tasks, using datasets ETTh1, ETTh2, Weather, and ILI.
    \item On cold-start forecasting, our methods have consistently achieved remarkable results, resulting in a higher discrepancy.
\end{itemize}

\section{Related Work}

\textbf{Basic Augmentation Methods:} As mentioned earlier, basic augmentation operations are scaling, flipping~\cite{basic}, window cropping or slicing~\cite{windowcrop}, Gaussian noise injection~\cite{gaussian}, dynamic time warping~\cite{survey}, and others. Scaling operations alter the magnitude of data within a defined range by using a randomly generated scalar multiplier, while flipping reverses the sequence of data points along the time axis~\cite{basic}. Window cropping is a method that selects random sections of a specific length from time series data, and each section is classified using a majority voting approach during testing. Gaussian noise injection is a technique that involves adding small amounts of noise or outliers to time series data while keeping the original labels intact~\cite{survey}. These data augmentation techniques are mostly suitable for time series classification or anomaly detection as they do not maintain temporal coherence, which is crucial for time series forecasting. When these methods are applied for forecasting, they yield inferior results~\cite{fr_aug}.

\textbf{Decomposition-based Augmentation Methods:} Some decomposition-based approaches break down time series data into constituent elements such as trend, seasonality, and residual components through techniques such as RobustSTL or STL. Subsequently, these elements are transformed and merged to generate new time series data by modifying weights and implementing statistical models~\cite{survey}. Some methods that rely on decomposition use Empirical mode decomposition (EMD), and one particular method is known as STAug method. The EMD algorithm breaks down a signal into multiple components called Intrinsic Mode Functions (IMFs). The initial IMFs capture high-frequency components, while the final IMFs represent low-frequency trend information. STAug utilizes two distinct instances, applies EMD to break them into components, and then reconstructs these subcomponents using randomly sampled weights from a uniform distribution. It then applies linear interpolation of these two weights to obtain augmented data for time series forecasting~\cite{staug}. We have conducted experiments using the STAug method, which serves as one of our baseline approaches.

\textbf{Frequency-Domain Augmentation Methods:} Some studies investigate data augmentation techniques that work on only the frequency domain; one of the most recent works was conducted by Gao et al.~\cite{freqdomain}. They use the Fourier transform to manipulate frequency components, such as adding perturbations or eliminating a few of them~\cite{survey}. One of the recent works is also FrAug methods, which consist of two methods: \textit{frequency masking} (FreqMask) and \textit{frequency mixing} (FreqMix)~\cite{fr_aug}. These techniques form the fundamental basis of our paper and serve as our baselines. FreqMask employs the Fast Fourier transform (FFT) to compute the frequency domain representations of input data. Subsequently, it generates augmented data by randomly masking specific frequency components and applies the inverse real FFT for signal reconstruction. The procedure for FreqMix is similar, except that it randomly substitutes the frequency components in one training instance with the frequency components of another training instance~\cite{fr_aug}. 

\section{Problem Formulation}

A multivariate time series sequence of length T and channels K is denoted as $\mathcal{X} = (x_{1},\dots,x_{T}) \in \mathbb{R}^{T \times K}$. As it is a \textbf{Time Series Forecasting Problem}, we only observe data up to timestamp $t<T$. Based on the forecasting horizon, we aim to forecast future values starting from $t+1,\dots, T$. Given input $X = (x_{1},\dots,x_{t}) \in \mathbb{R}^{t \times K}$, the objective is to learn a model $f$ that forecasts $F = (x_{t+1},\dots,x_{T}) \in \mathbb{R}^{(T-t) \times K}$ to obtain predictions for the future part. We learn a model for TSF as:

$$ f_{\theta}:\mathbb{R}^{t \times K} \rightarrow  \mathbb{R}^{(T-t) \times K} \hspace{0.5mm}, \hspace{2.5mm} X \mapsto f_{\theta}(X)$$

The data $X$ may be limited or insufficient for time series forecasting, the synthetic time series data $\overline X = (\overline x_{1},\dots, \overline x_{t}) \in \mathbb{R}^{t \times K}$ is generated by applying data augmentation methods. Therefore, we want to learn a model:

$$ f_{\theta}: \mathbb{R}^{t \times K} \times \mathbb{R}^{t \times K} \rightarrow  \mathbb{R}^{(T-t) \times K} \hspace{0.5mm}, \hspace{2.5mm} X' \mapsto f_{\theta}(X')$$
where the data: 
$$X' = [X, \overline X] = \big[[x_{1},\dots,x_{t}], [\overline x_{1},\dots, \overline x_{t}]\big]$$.

The loss functions are the mean squared error (MSE) and the mean absolute error (MAE) averaged over future time points.

\section{Methodology}
This section provides a detailed structure of how the training framework works with our proposed methods. We will also detail these two simple methods with illustrations and pseudocodes.

\subsection{Overview}
\begin{figure}[h!]
\centering
\includegraphics[width=1.0\textwidth, clip = true]{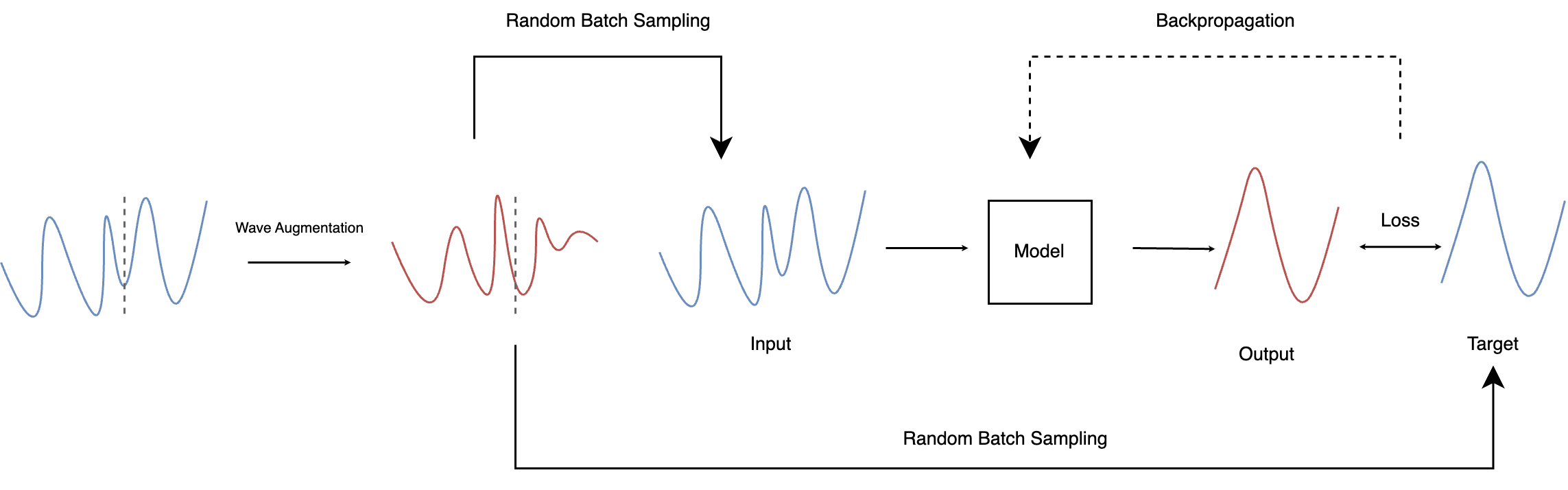}
\caption{Wavelet-Integrated Training Framework} \label{fig1}
\end{figure}

The discrete wavelet transform (DWT), as discussed in Appendix A, is proposed for enhancing time series data by preserving intricate data characteristics and facilitating multi-resolution analysis that accounts for various frequencies at different resolutions. The illustration in Fig.~\ref{fig1} depicts a framework of the training stages incorporating wavelet augmentations, which involve the concatenation of the look-back window and the forecasting horizon prior to transformation and augmentation. Batch sampling of the generated synthetic data is conducted according to a predefined hyperparameter called the \textit{sampling rate}. These batches are subsequently used to split the data into the look-back window and target horizon, after which they are concatenated with the original data. By maintaining consistency with the notations presented in the Problem Formulation section, the random batch sampling operation can be defined as:

$$ f_{\text{sample}}: \mathbb{R}^{b \times t \times K} \rightarrow \mathbb{R}^{n \times t \times K} \hspace{0.5mm}, \hspace{2.5mm} \overline X \mapsto f_{\text{sample}}(\overline X) $$ where $b$ is our initial batch size and $n$ is determined based on the hyperparameter \textit{sampling rate} with $n \leq b$.

Our selected model, DLinear~\cite{dlinear}, is characterized by a decomposition scheme and a linear layers. Given the look-back window as $X \in \mathbb{R}^{L \times K}$, DLinear model initially decomposes it into a trend component $X_{t}\in \mathbb{R}^{L \times K}$ and a residual part $X_{r} \in \mathbb{R}^{L \times K}$. In order to produce the forecasting output $F \in \mathbb{R}^{T \times K}$, linear layers are employed for both the trend and residual components as follows: $H_t = W_t X_t$ where $W_t \in \mathbb{R}^{T \times L}$ and $H_r = W_r X_r$ where $W_r \in \mathbb{R}^{T \times L}$ respectively. The forecast output is obtained by summing both components $H_t$ and $H_r$~\cite{dlinear}. Augmentations are exclusively implemented during the training phase, while the original test samples remain unaltered for evaluation purposes.

\subsection{Wavelet Masking (WaveMask)}

\begin{figure}[h!]
\centering
\includegraphics[width=0.95\linewidth, clip=true]{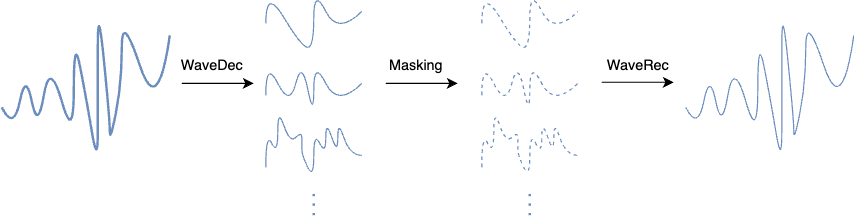}
\caption{The proposed wavelet masking (WaveMask) augmentation method} 
\label{fig2}
\end{figure}

Our first method, \textbf{WaveMask}, creates a synthetic dataset based on the input, which is a concatenation of the look-back window and the target horizon (Line 1 in Alg.~\ref{wavemask}). The method is illustrated in Fig.~\ref{fig2}. WaveDec (Wavelet Decomposition) performs multilevel discrete wavelet transform (DWT) on the input signal. It decomposes the signal into the last approximation and all detail coefficients for each resolution level $\mathcal{L}$ (Line 3 in Alg.~\ref{wavemask}). We mask the wavelet coefficients based on the array of hyperparameter \textit{rates} ($\mu^{\mathcal{L} +1}$), which contains parameter \textit{rate} for the last approximation and all detail coefficients; it has a length of $\mathcal{L} + 1$ (Line 4 in Alg.~\ref{wavemask}). It determines the probability of each coefficient being masked (set to zero) at that specific level, allowing for level-specific control over the masking process. $\text{CreateRandomMask}$ generates boolean values (True or False) to determine whether to mask or not mask sections according to the \textit{rate}, and the $\text{Masking}$ function replaces True elements with zeros (Line 5 \& 6 in Alg.~\ref{wavemask}). WaveRec (Wavelet Reconstruction) applies the inverse discrete wavelet transform (IDWT) to reconstruct the signal from its wavelet coefficients. It synthesizes the original signal structure from the modified/masked wavelet coefficients, effectively reversing the decomposition process (Line 8 in Alg.~\ref{wavemask}). We implemented it for each channel (c) separately (Line 2 in Alg.~\ref{wavemask}). Masking wavelet coefficients allow us to eliminate frequency components at various time scales selectively. When we reconstruct the signal using the masked coefficients, we obtain a modified version of the original signal in which specific frequency components have been eliminated.

\begin{algorithm}[H]
\footnotesize
\SetAlgoLined
\KwIn{\textbf{Look-back window} $x$, \textbf{target horizon} $y$, \textbf{resolution level} $\mathcal{L}$, \textbf{mask rate} $\mu^{\mathcal{L}+1}$}
\KwOut{Masked sequence $\Tilde{s}$}

$s \leftarrow x || y$ \tcp*{Concatenate $x$ and $y$}
\For{\textbf{each} channel $c$}{
    $W^{\mathcal{L}+1} \leftarrow \text{WaveDec}(s[:,:,c])$ \tcp*{Multilevel decomposition}
    \For{$i \in \{1, \ldots, \mathcal{L}+1\}$}{
        $m \leftarrow \text{CreateRandomMask}(\text{len}(W[i]), \mu[i])$ 
        
        $\Tilde{W}[i] \leftarrow \text{Masking}(W[i], m)$ \tcp*{Apply mask to coefficients}
    }
    $\Tilde{s}[{\text{c}}] \leftarrow \text{WaveRec} (\Tilde{W})$ \tcp*{Multilevel reconstruction}
}
\Return{$\Tilde{s}$}
\caption{WaveMask}
\label{wavemask}
\end{algorithm}

\subsection{Wavelet Mixing (WaveMix)}

\begin{figure}[h!]
\centering
\includegraphics[width=0.8\textwidth, clip=true]{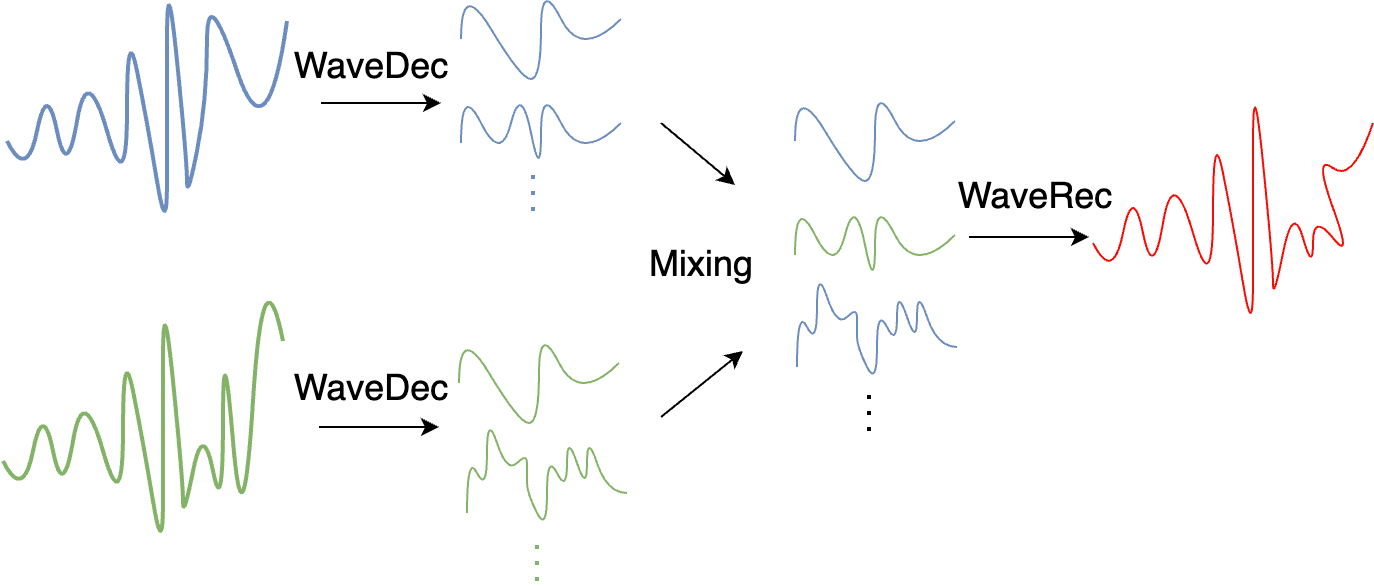}
\caption{The proposed wavelet mixing (WaveMix) augmentation method} \label{fig3}
\end{figure}

The second method, \textbf{WaveMix},  extends the \textbf{WaveMask} concept to mix two input signals using wavelet transformation. It takes two different input pairs, concatenates these look-back windows with the corresponding target horizons, and outputs augmented data after applying the transformation pipeline. The pipeline is illustrated in Fig.~\ref{fig3}. In the same way as our first method, WaveMask, we implement wavelet decomposition for both input pairs separately. $\text{CreateRandomMask}$  generates a binary mask based on the mix rate $\mu[i]$ ($i = 1, \ldots, \mathcal{L}+1$) for each wavelet level, determining which coefficients to select from each input signal. $\text{BitwiseNOT}$ operation creates a complementary mask, ensuring that coefficients not selected from one signal are taken from the other.  $\text{Masking}$ applies the generated masks to the wavelet coefficients of each input signal and we add both to select portions from each effectively (Lines 5-7 in Alg.~\ref{wavemask}). We apply the same procedure as wavelet reconstruction to obtain the augmented data. By exchanging wavelet coefficients between inputs, the frequency components of one data instance are integrated into the other, and vice versa. This process enables the mixing of frequency information across different time granularities between the two data instances. We ran the process for each channel separately.

\begin{algorithm}[H]
\footnotesize
\SetAlgoLined
\KwIn{\textbf{Look-back window} $x_1$, \textbf{target horizon} $y_1$, \textbf{another training sample pair} $x_2$, $y_2$, \textbf{resolution level} $\mathcal{L}$, \textbf{mix rate} $\mu^{\mathcal{L}+1}$}
\KwOut{Masked sequence $\Tilde{s}$}

$s_1 \leftarrow x_1 || y_1$; $s_2 \leftarrow x_2 || y_2$ \tcp*{Concatenate $x$ and $y$}

\For{\textbf{each} channel $c$}{
 $W_1^{\mathcal{L}+1} \leftarrow \text{WaveDec}(s_1[:, :, c])$; $W_2^{\mathcal{L}+1} \leftarrow \text{WaveDec}(s_2[:, :, c])$;
 
    \For{$i \in \{1, \ldots, \mathcal{L}+1\}$}{
        $m_1 \leftarrow \text{CreateRandomMask}(\text{len}(W_1[i]), \mu[i])$ ;
        
        $m_2 \leftarrow \text{BitwiseNOT}(m_1)$;
        
        $\Tilde{W}[i] \leftarrow \text{Masking}(W_1[i], m_1) + \text{Masking}(W_2[i], m_2) $ ;
    }
    $\Tilde{s}[c] \leftarrow \text{WaveRec} (\Tilde{W})$ \tcp*{Multilevel reconstruction}
}

\Return{$\Tilde{s}$}
\caption{WaveMix}
\label{wavemix}
\end{algorithm}
\section{Experiments}

The first section provides an overview of the datasets, their statistical information, the baseline methods, and the experimental setup. Section 5.2 presents the main results regarding four datasets. The final section presents the ablation study of cold-start forecasting, wherein models are trained on down-sampled training datasets.

\subsection{Datasets, Baselines and Experimental Setup}

Our methods are assessed using four datasets: ETTh1, ETTh2, Weather, and ILI~\cite{dataset1,dataset2}. We selected these datasets as they represent a diverse range of time series forecasting tasks across different domains. The statistical details for the datasets can be found in Table~\ref{tab:tb1}. Our approach follows the methodology delineated in the prior research~\cite{fr_aug,dlinear}, which utilized a 24-length look-back window and forecasting horizons of (24, 36, 48, 60) for ILI. For the rest, a 336-length look-back window and forecasting horizons of (96, 192, 336, 720) were used. The forecasting horizons were chosen to evaluate performance on short-, medium-, and long-term predictions, aligning with common practices in time series forecasting literature~\cite{survey}. The datasets are segmented into train, validation, and test sets at a ratio of 6:2:2. The comparison includes the FreqMask, FreqMix~\cite{fr_aug}, and STAug~\cite{staug} techniques. We implemented an optimized version of STAug that performs Empirical Mode Decomposition (EMD) for the entire sequence. The evaluation employs the following metrics:

\begin{itemize}[label=$\bullet$] \item Mean Squared Error (MSE) - In the context of multivariate time series forecasting with multiple channels, MSE is calculated by squaring the differences between actual and predicted values for each channel, averaging these squared errors across all data points, and then averaging the results across all channels. \item Mean Absolute Error (MAE) - Similar to MSE, MAE is derived by taking the absolute differences between actual and predicted values for each channel, averaging these absolute errors across all data points, and then averaging the results across all channels. \end{itemize}

We utilized DLinear model~\cite{dlinear} and treated each forecasting horizon as a distinct task. Our approaches fine-tune the parameters for each task through Random Search, while the baseline methods employ Grid Search~\cite{search}. The optimized hyperparameters are provided in Appendix B. In contrast to the experimental setup of the baseline methods, we executed a 30-epoch run with a patience level of 12 and selected the optimal one based on validation loss. The batch size is configured to 64, and a learning schedule is incorporated. A random sampling strategy is utilized to choose a subset of the generated data.

\begin{table}[h!]
\caption{The statistical information of datasets}
\label{tab:tb1}
\renewcommand{\arraystretch}{1.0}
\setlength{\tabcolsep}{8pt}
\centering
\small
\begin{tabular}{c|c|c|c|c}
\hline
Datasets & ETTh1 & ETTh2 & Weather & ILI \\ \hline
Variates & 7 & 7 & 21 & 7 \\
Frequency & 1 hour & 1 hour & 10 min & 1 week \\
Total Timesteps & 17420 & 17420 & 52696 & 966 \\ \hline
\end{tabular}%

\end{table}

\subsection{Main Results}

Table~\ref{tab:tb2} and Table~\ref{tab:tb3} present comparisons between our methods and baselines in terms of the metrics Mean Squared Error (MSE) and Mean Absolute Error (MAE). The best result is indicated in bold, while the second most favorable outcome is underlined. We conduct the experiments a total of 10 times and report standard deviations in contrast to the experiments conducted by the previous studies~\cite{fr_aug,staug}. Every forecasting horizon is treated as an individual task, and the parameters are optimized independently. Among the 16 tasks, our methods, WaveMask and WaveMix, outperform all baselines in 12. Among these four tasks, they rank as the second most effective methods. The STAug in the Weather dataset exceeds the memory capacity of the GPU due to creating a tensor-based on various factors such as data length, maximum Intrinsic Mode Function (IMF), sequence length, forecasting horizon, and channels. We can observe that STAug method is not stable and standard deviations are high, especially in the ETTh2 dataset.

\begin{table}[h!]
\caption{Comparison of augmentation methods for various forecasting horizons under \textbf{MSE}. * denotes out-of-memory issue}
\label{tab:tb2}
\renewcommand{\arraystretch}{2.0}
\setlength{\tabcolsep}{10pt} 
\resizebox{\textwidth}{!}{%
\begin{tabular}{clcccc|cc}
\hline
\multicolumn{2}{c}{Methods} & \multicolumn{1}{c|}{None} & \multicolumn{1}{c|}{FreqMask} & \multicolumn{1}{c|}{FreqMix} & STAug & \multicolumn{1}{c|}{WaveMask} & WaveMix \\ \hline
\multicolumn{1}{c|}{\multirow{4}{*}{ETTh1}} & \multicolumn{1}{l|}{96} & 0.3708 ± 1.66e-05 & 0.3698 ± 7.69e-05 & 0.3698 ± 0.0001 & 0.3707 ± 0.007 & {\ul 0.3698 ± 2.07e-05} & \textbf{0.3696 ± 1.58e-05} \\
\multicolumn{1}{c|}{} & \multicolumn{1}{l|}{192} & 0.453 ± 0.0002 & 0.4559 ± 0.0008 & 0.4537 ± 0.0004 & 0.4552 ± 0.06 & {\ul 0.4484 ± 0.0002} & \textbf{0.4464 ± 0.0003} \\
\multicolumn{1}{c|}{} & \multicolumn{1}{l|}{336} & 0.5382 ± 0.0007 & 0.5170 ± 0.002 & 0.5307 ± 0.01 & 0.5199 ± 0.01 & {\ul 0.4995 ± 0.0002} & \textbf{0.4967 ± 0.0001} \\
\multicolumn{1}{c|}{} & \multicolumn{1}{l|}{720} & 0.5109 ± 0.0002 & 0.5106 ± 0.002 & \textbf{0.4981 ± 0.0002} & {\ul 0.4866 ± 0.02} & 0.5089 ± 0.0002 & 0.5076 ± 0.0002 \\ \hline
\multicolumn{1}{c|}{\multirow{4}{*}{ETTh2}} & \multicolumn{1}{l|}{96} & 0.3035 ± 3.69e-05 & 0.3096 ± 0.003 & 0.3031 ± 0.0003 & 0.4179 ± 0.03 & \textbf{0.3018 ± 7.76e-05} & {\ul 0.303 ± 0.0004} \\
\multicolumn{1}{c|}{} & \multicolumn{1}{l|}{192} & 0.4171 ± 0.002 & 0.4153 ± 0.07 & {\ul 0.4152 ± 0.003} & 0.5778 ± 0.13 & \textbf{0.4152 ± 0.002} & 0.4230 ± 0.005 \\
\multicolumn{1}{c|}{} & \multicolumn{1}{l|}{336} & 0.4799 ± 0.003 & \textbf{0.4512 ± 0.005} & 0.4764 ± 0.002 & 1.1409 ± 0.23 & 0.47 ± 0.003 & {\ul 0.4644 ± 0.003} \\
\multicolumn{1}{c|}{} & \multicolumn{1}{l|}{720} & 0.5471 ± 0.0004 & {\ul 0.54 ± 0.003} & 0.5473 ± 0.0007 & 1.9252 ± 0.28 & 0.5469 ± 0.0004 & \textbf{0.5117 ± 0.0007} \\ \hline
\multicolumn{1}{c|}{\multirow{4}{*}{Weather}} & \multicolumn{1}{l|}{96} & 0.l858 ± 2.27e-06 & 0.1879 ± 0.002 & 0.1895 ± 0.002 & * & {\ul 0.1855 ± 1.34e-05} & \textbf{0.1852 ± 2.42e-05} \\
\multicolumn{1}{c|}{} & \multicolumn{1}{l|}{192} & 0.2389 ± 8.16e-07 & 0.2382 ± 0.0002 & 0.2393 ± 5.8e-05 & * & \textbf{0.2352 ± 3.02e-05} & {\ul 0.2353 ± 4.21e-05} \\
\multicolumn{1}{c|}{} & \multicolumn{1}{l|}{336} & 0.2921 ± 5.49e-07 & 0.2908 ± 0.0002 & 0.2922 ± 3.1e-05 & * & \textbf{0.2893 ± 5.34e-07} & {\ul 0.2896 ± 5.69e-05} \\
\multicolumn{1}{c|}{} & \multicolumn{1}{l|}{720} & 0.3555 ± 2.18e-07 & \textbf{0.3505 ± 0.0002} & 0.3553 ± 4.33e-05 & * & 0.3553 ± 6.45e-07 & {\ul 0.3515 ± 4.04e-05} \\ \hline
\multicolumn{1}{c|}{\multirow{4}{*}{ILI}} & \multicolumn{1}{c|}{24} & 2.7759 ± 0.008 & 2.7855 ± 0.01 & 2.7797 ± 0.008 & \textbf{2.7385 ± 0.02} & 2.7768 ± 0.009 & {\ul 2.775 ± 0.008} \\
\multicolumn{1}{c|}{} & \multicolumn{1}{c|}{36} & 3.0018 ± 0.01 & 3.053± 0.008 & 3.034 ± 0.01 & 3.0627 ± 0.03 & {\ul 3.0014 ± 0.008} & \textbf{2.8912 ± 0.003} \\
\multicolumn{1}{c|}{} & \multicolumn{1}{c|}{48} & 2.9205 ± 0.009 & 2.9553 ± 0.009 & 2.9428 ± 0.008 & 2.9598 ± 0.04 & {\ul 2.9082 ± 0.008} & \textbf{2.819 ± 0.001} \\
\multicolumn{1}{c|}{} & \multicolumn{1}{c|}{60} & 3.0579 ± 0.006 & 3.0576 ± 0.007 & 3.0595 ± 0.007 & 3.1042 ± 0.02 & {\ul 3.0322 ± 0.005} & \textbf{3.0195 ± 0.004} \\ \cline{2-8} 
\end{tabular}%
}
\end{table}

\begin{table}[h!]
\caption{Comparison of augmentation methods for various forecasting horizons under \textbf{MAE}. * denotes out-of-memory issue}
\label{tab:tb3}
\renewcommand{\arraystretch}{2.0}
\setlength{\tabcolsep}{10pt}
\resizebox{\textwidth}{!}{%
\begin{tabular}{clcccc|cc}
\hline
\multicolumn{2}{c}{Methods} & \multicolumn{1}{c|}{None} & \multicolumn{1}{c|}{FreqMask} & \multicolumn{1}{c|}{FreqMix} & STAug & \multicolumn{1}{c|}{WaveMask} & WaveMix \\ \hline
\multicolumn{1}{c|}{\multirow{4}{*}{ETTh1}} & \multicolumn{1}{l|}{96} & 0.3946 ± 2.06e-05 & 0.3942 ± 0.0001 & 0.3947 ± 0.0002 & 0.3992 ± 0.008 & {\ul 0.3939 ± 2.96e-05} & \textbf{0.3936 ± 1.69e-05} \\
\multicolumn{1}{c|}{} & \multicolumn{1}{l|}{192} & 0.4660 ± 0.0002 & 0.4682 ± 0.0007 & 0.4667 ± 0.0004 & 0.4645 ± 0.003 & {\ul 0.4619 ± 0.0002} & \textbf{0.4601 ± 0.0003} \\
\multicolumn{1}{c|}{} & \multicolumn{1}{l|}{336} & 0.5257 ± 0.0004 & 0.5112 ± 0.001 & 0.5163 ± 0.01 & 0.5064 ± 0.008 & {\ul 0.4967 ± 0.0001} & \textbf{0.4944 ± 0.0001} \\
\multicolumn{1}{c|}{} & \multicolumn{1}{l|}{720} & 0.5229 ± 0.0001 & 0.5209 ± 0.0008 & \textbf{0.5105 ± 0.0002} & 0.5164 ± 0.02 & 0.5212 ± 0.0001 & {\ul 0.5189 ± 0.0001} \\ \hline
\multicolumn{1}{c|}{\multirow{4}{*}{ETTh2}} & \multicolumn{1}{l|}{96} & 0.3652 ± 3.84e-05 & 0.3646 ± 0.002 & {\ul 0.3648 ± 0.0003} & 0.4393 ± 0.02 & \textbf{0.3641 ± 6.87e-05} & 0.3650 ± 0.0004 \\
\multicolumn{1}{c|}{} & \multicolumn{1}{l|}{192} & 0.4317± 0.001 & 0.4311 ± 0.005 & {\ul 0.4309 ± 0.002} & 0.5179 ± 0.07 & \textbf{0.4296 ± 0.002} & 0.4367 ± 0.003 \\
\multicolumn{1}{c|}{} & \multicolumn{1}{l|}{336} & 0.479 ± 0.002 & \textbf{0.4638 ± 0.003} & 0.4771 ± 0.001 & 0.7465 ± 0.09 & 0.4735 ± 0.002 & {\ul 0.4701 ± 0.002} \\
\multicolumn{1}{c|}{} & \multicolumn{1}{l|}{720} & 0.5276 ± 0.0002 & {\ul 0.5239 ± 0.002} & 0.5278 ± 0.0003 & 1.0115 ± 0.08 & 0.5274 ± 0.0002 & \textbf{0.5090 ± 0.0004} \\ \hline
\multicolumn{1}{c|}{\multirow{4}{*}{Weather}} & \multicolumn{1}{l|}{96} & 0.2595 ± 5.24e-06 & 0.2629 ± 0.004 & 0.2662 ± 0.004 & * & {\ul 0.2590 ± 2.87e-05} & \textbf{0.2584 ± 5.7e-05} \\
\multicolumn{1}{c|}{} & \multicolumn{1}{l|}{192} & 0.3138 ± 3.84e-06 & 0.3127 ± 0.0003 & 0.3144 ± 8.08e-05 & * & \textbf{0.3090 ± 4.47e-05} & {\ul 0.3094 ± 5.5e-05} \\
\multicolumn{1}{c|}{} & \multicolumn{1}{l|}{336} & 0.3583 ± 2.64e-06 & 0.3569 ± 0.0002 & 0.3586 ± 3.90e-05 & * & \textbf{0.3542 ± 3.41e-06} & {\ul 0.3556 ± 7.57e-05} \\
\multicolumn{1}{c|}{} & \multicolumn{1}{l|}{720} & 0.4079 ± 5.42e-07 & \textbf{0.4022 ± 0.0002} & 0.4082 ± 6.19e-05 & * & 0.4077 ± 9.80e-07 & {\ul 0.4039 ± 4.78e-05} \\ \hline
\multicolumn{1}{c|}{\multirow{4}{*}{ILI}} & \multicolumn{1}{c|}{24} & 1.1371 ± 0.002 & 1.1385 ± 0.003 & 1.1373 ± 0.002 & {\ul 1.1333 ± 0.005} & 1.1354 ± 0.002 & \textbf{1.1333 ± 0.002} \\
\multicolumn{1}{c|}{} & \multicolumn{1}{c|}{36} & 1.1604 ± 0.003 & 1.1710 ± 0.002 & 1.1678 ± 0.003 & 1.1781 ± 0.007 & {\ul 1.1583 ± 0.002} & \textbf{1.1263 ± 0.0008} \\
\multicolumn{1}{c|}{} & \multicolumn{1}{c|}{48} & 1.1575 ± 0.002 & 1.1604 ± 0.002 & 1.1601 ± 0.002 & 1.1668 ± 0.006 & {\ul 1.1531 ± 0.002} & \textbf{1.1256 ± 0.0002} \\
\multicolumn{1}{c|}{} & \multicolumn{1}{c|}{60} & 1.2029 ± 0.001 & 1.1995 ± 0.002 & 1.2 ± 0.002 & 1.2110 ± 0.006 & {\ul 1.1953 ± 0.001} & \textbf{1.1880 ± 0.0009} \\ \cline{2-8} 
\end{tabular}%
}
\end{table}

\subsection{Ablation Study}

\begin{figure}[h!]
    \centering
    \begin{minipage}{0.5\textwidth}  % Adjust width to accommodate images
        \centering
        \includegraphics[width=\linewidth,height=3cm]{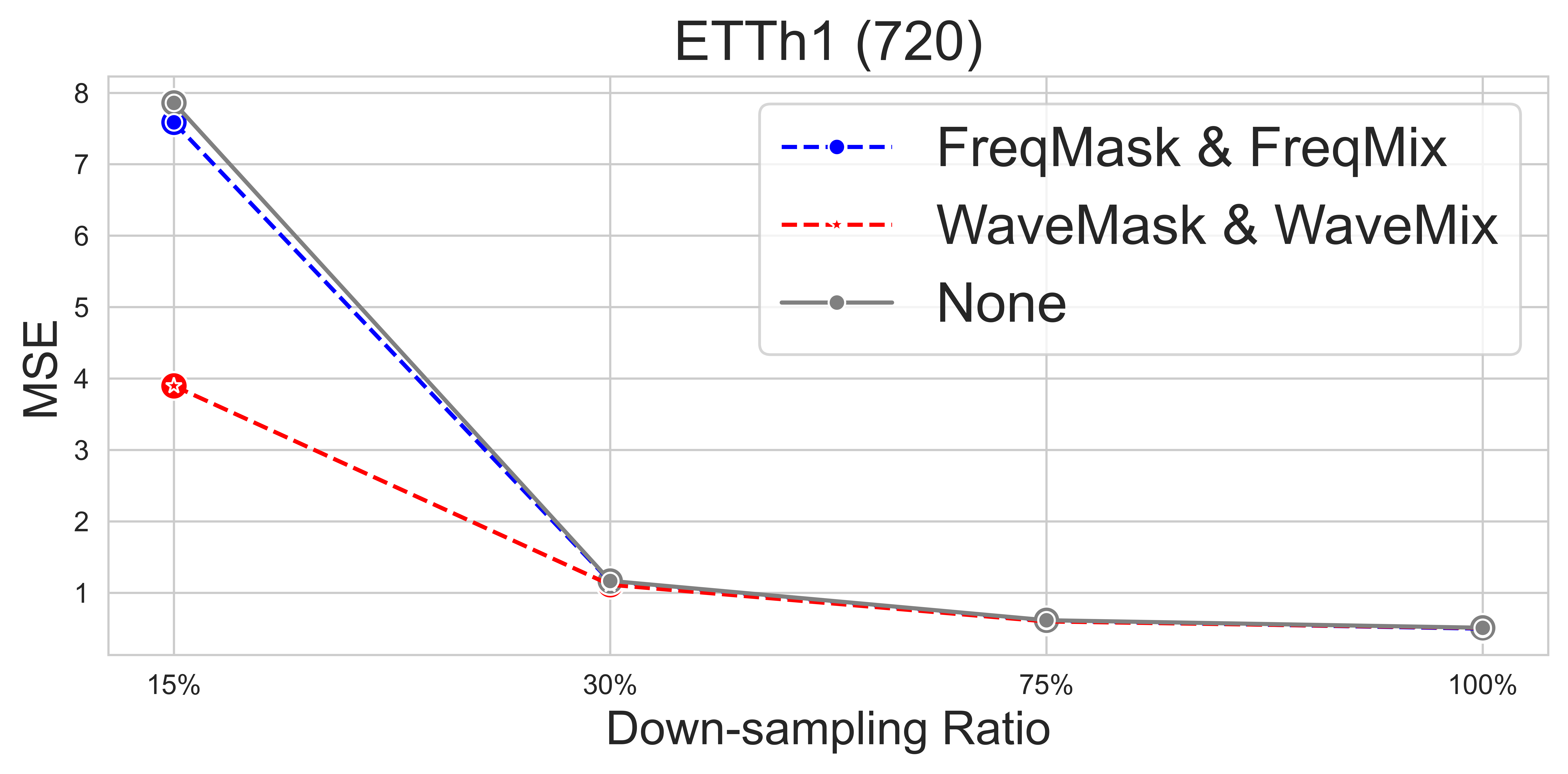}
    \end{minipage}\hfill
    \begin{minipage}{0.5\textwidth}  % Adjust width to accommodate images
        \centering
        \includegraphics[width=\linewidth,height=3cm]{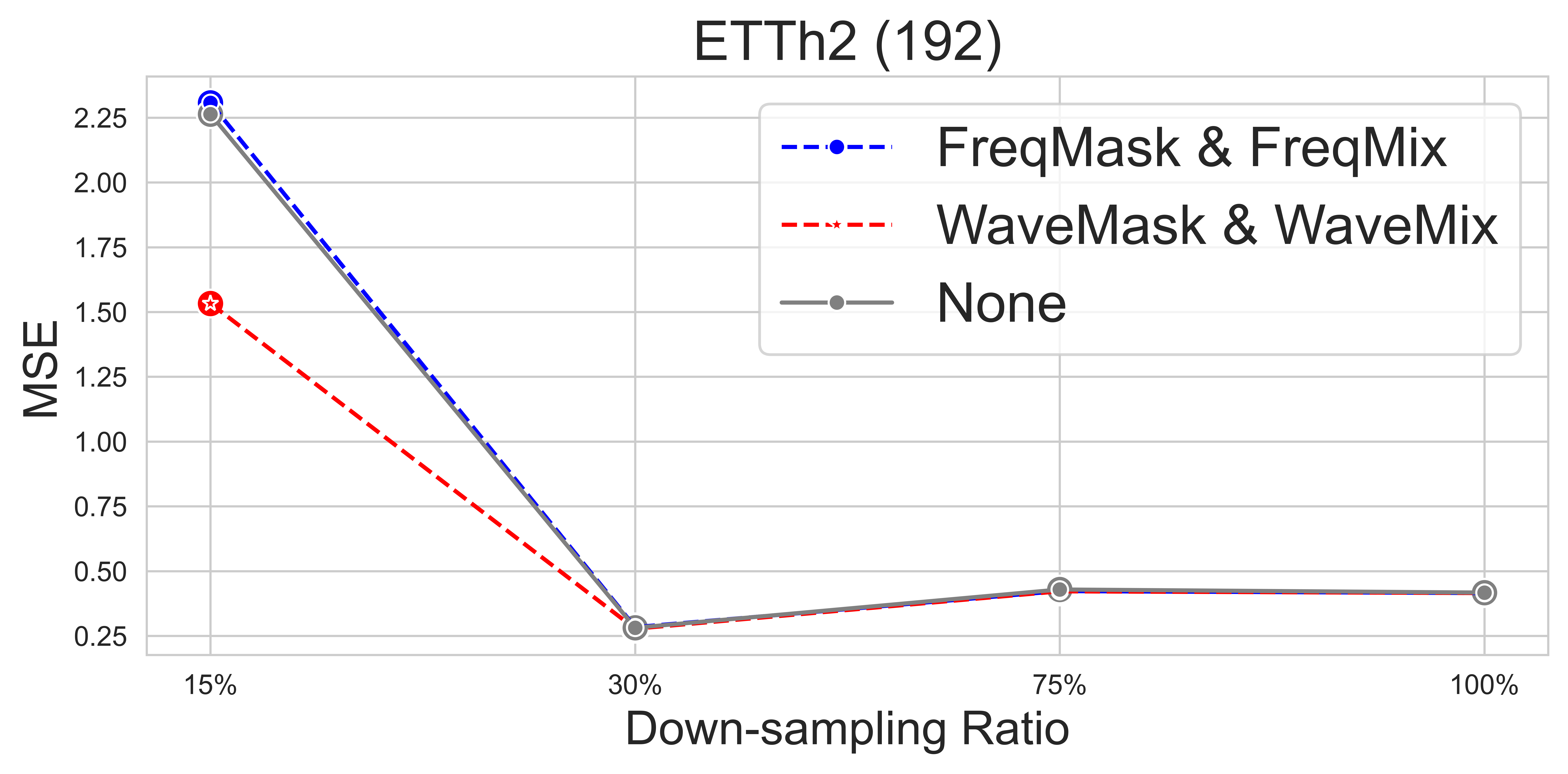}
    \end{minipage}\hfill
    
    \begin{minipage}{0.5\textwidth}  % Adjust width to accommodate images
        \centering
        \includegraphics[width=\linewidth,height=3cm]{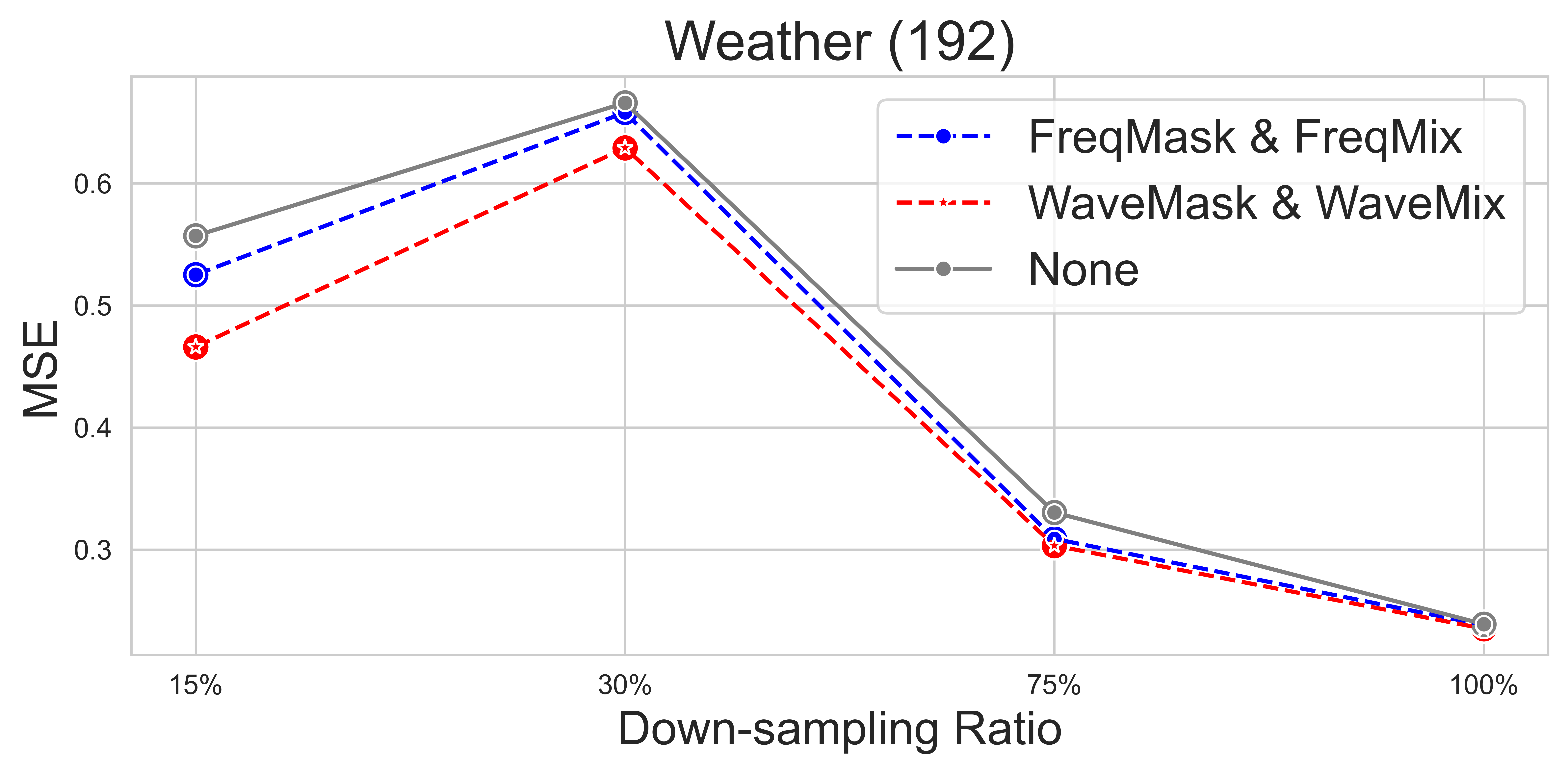}
    \end{minipage}\hfill
    \begin{minipage}{0.5\textwidth}  % Adjust width to accommodate images
        \centering
        \includegraphics[width=\linewidth,height=3cm]{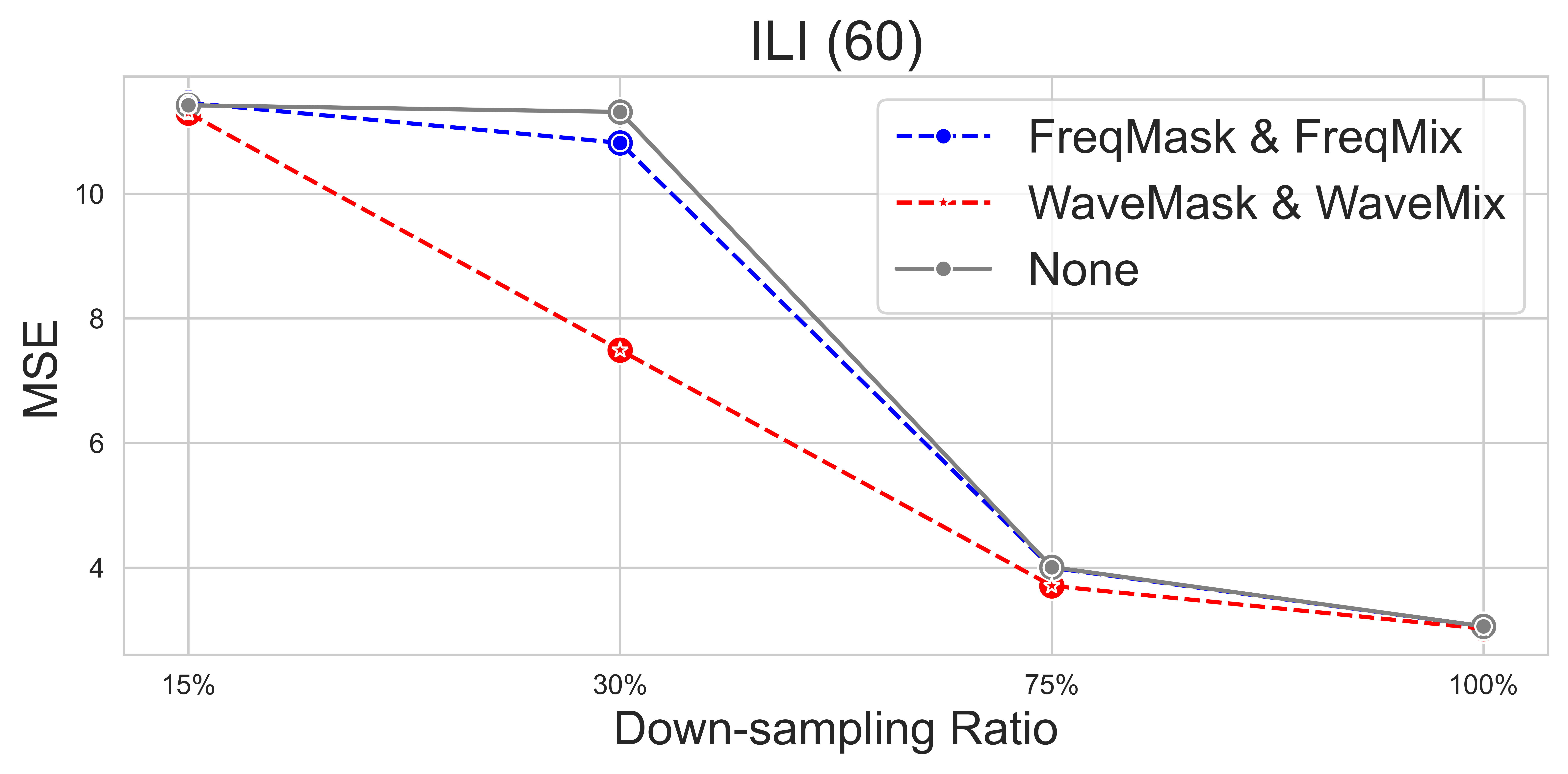}
    \end{minipage}
    \caption{Cold-start forecasting for different datasets. Parentheses show the forecasting horizon}
    \label{fig:ablation}
\end{figure}

We carried out an evaluation of the cold-start forecasting task utilizing the ETTh1, ETTh2, Weather, and ILI datasets across different forecasting horizons. The downsampling rates indicate the percentage of the training dataset that was retained while keeping the test dataset unchanged. We chose different down-sampling rates of 15\%, 30\%, and 75\% for the evaluations. The performance, as depicted in Fig.~\ref{fig:ablation}, is presented using the Mean Squared Error (MSE) metric, with the forecasting horizon specified in parentheses. The optimal performance, indicated by the minimum value of MSE values, was chosen from the FreqMask \& FreqMix and WaveMask \& WaveMix combinations, respectively. It is obvious that our methods deliver superior outcomes, particularly noticeable when the downsampling rate is as low as 15\%. Furthermore, it is noteworthy that we did not fine-tune hyperparameters for the augmentation techniques; instead, we utilized the same parameters optimized on the full training data. We hypothesize that our methods, WaveMask and WaveMix, may generate exceptional results at lower downsampling rates through hyperparameter optimization, and this area of research holds potential for future investigation.  

\section{Discussion \& Conclusion}

All of these data augmentation methods can be assessed across a wide range of diverse dataset domains, in addition to the experiments already conducted. To the best of our knowledge, there are no state-of-the-art techniques other than those we evaluated, except TimeGAN~\cite{timegan}, to compare with our methods as the methods we used are also evaluated with little data in cold-start forecasting setting. The evaluations cannot be directly compared to the paper~\cite{fr_aug} because we conducted them using 30 epochs instead of 10 epochs and selected the best model from the validation dataset. In addition, we performed the running task 10 times and reported the standard deviation. Instead of utilizing Random Search, more sophisticated search techniques could be employed to discover optimal parameters for WaveMask and WaveMix that may result in exceptional outcomes. We initially employed DLinear model~\cite{dlinear}, however, for further progress, TSMixer~\cite{tsmixer} can be utilized as it demonstrates superior outcomes and attains state-of-the-art performance. Furthermore, the augmentation methods can be implemented following linear projection layers and techniques that function on representations of time series, such as in TS2Vec~\cite{ts2vec}.

In conclusion, both our methods, WaveMask and WaveMix, outperform the baseline methods~\cite{fr_aug,staug} and are more straightforward and comprehensive techniques compared to methods that solely focus on the frequency domain without considering the time domain. Based on the ablation study, our methods have also been shown to achieve outstanding results in cold-start forecasting.

%
% ---- Bibliography ----
%
% BibTeX users should specify bibliography style 'splncs04'.
% References will then be sorted and formatted in the correct style.
%
%\bibliographystyle{splncs04}

\appendix

\section{Discrete Wavelet Transform (DWT)}

During a wavelet transform, a signal is subjected to multiplication by a wavelet function, which is a localized wave with finite energy. The resulting transform is then analyzed for each segment. The equation that defines a continuous wavelet transform (CWT) is as follows:

\begin{equation}
    H(x)=\frac{1}{|\sqrt{\zeta}|} \int x(t) \cdot \psi^*\left(\frac{t-\tau}{\zeta}\right) dt
    \label{eq: cwt}
\end{equation}
where $H(x)$ is the wavelet transform for the signal $x(t)$ as a function of time $t$, $\zeta$ is the scale parameter, $\tau$ is the time parameter, and $\psi$ is the mother wavelet or the basis function. High frequencies, or small scales, compress the signal and convey global information, while low frequencies, or large scales, expand the signal and reveal hidden detailed information within it. Nevertheless, the cost remains high, and DWT offers effective computation through subband coding, where the signal undergoes filtering with distinct cutoff frequencies at various scales. The DWT is calculated sequentially by applying high-pass and low-pass filters to a signal, resulting in detail and approximation coefficients. The half-band filters reduce the signal's sampling rate by a factor of 2 at each level of decomposition. This method of breaking down and refining can be iterated until the desired level has been achieved. The original signal can be reconstructed by up-sampling the approximation and detail coefficients at each level, filtering them through high- and low-pass synthesis filters, and adding them together~\cite{dwtexp,dwtexp3}.

\section{Hyperparameters}

Random Search is applied to optimize the hyperparameters for our methods, WaveMask and WaveMix, while Grid Search is used for the remaining methods. The parameters that have been optimized are provided below:

\begin{table}[h!]
\caption{Hyperparameters for each dataset}
\label{tab:parameters}
\centering
\renewcommand{\arraystretch}{2.5}
\setlength{\tabcolsep}{4pt} 
\resizebox{\textwidth}{!}{%
\begin{tabular}{c|c|cccc|cccc|c|c|cc}
\hline
Datasets &  & \multicolumn{4}{c|}{WaveMask} & \multicolumn{4}{c|}{WaveMix} & \multicolumn{1}{l|}{FreqMask} & \multicolumn{1}{l|}{FreqMix} & \multicolumn{2}{c}{STAug} \\
 &  & Mother Wavelet & Level & Rates & Sampling rate & Mother Wavelet & Level & Rates & Sampling rate & Rate & Rate & Rate & n\_IMF \\ \hline
\multirow{4}{*}{ETTh1} & 96 & db2 & 3 & {[}0.5, 0.3, 0.9, 0.9{]} & 0.2 & db3 & 1 & {[}0.0, 0.9{]} & 0.2 & 0.1 & 0.2 & 0.9 & 100 \\
 & 192 & db3 & 1 & {[}0.0, 1.0{]} & 0.2 & db3 & 1 & {[}1.0, 0.4{]} & 0.8 & 0.5 & 0.1 & 0.9 & 900 \\
 & 336 & db25 & 1 & {[}0.1, 0.9{]} & 0.8 & db3 & 1 & {[}0.0, 0.9{]} & 0.8 & 0.5 & 0.1 & 0.8 & 200 \\
 & 720 & db1 & 1 & {[}0.4, 0.9{]} & 0.2 & db5 & 1 & {[}0.1, 0.9{]} & 0.8 & 0.4 & 0.6 & 0.7 & 1000 \\ \hline
\multirow{4}{*}{ETTh2} & 96 & db26 & 2 & {[}0.4, 0.9, 0.0{]} & 0.5 & db25 & 2 & {[}0.9, 0.9, 0.1{]} & 0.2 & 0.6 & 0.9 & 0.4 & 2000 \\
 & 192 & db26 & 2 & {[}0.6, 0.7, 0.5{]} & 0.8 & db1 & 3 & {[}0.9, 0.4, 0.1, 0.8{]} & 0.5 & 0.6 & 0.8 & 0.9 & 200 \\
 & 336 & db1 & 3 & {[}0.2, 0.7, 0.9, 0.4{]} & 0.8 & db25 & 3 & {[}0.9, 0.1, 0.2, 0.5{]} & 0.8 & 0.1 & 0.8 & 0.6 & 100 \\
 & 720 & db5 & 4 & {[}0.8, 0.9, 0.4, 0.9, 0.4{]} & 0.2 & db5 & 1 & {[}0.5, 0.1{]} & 1.0 & 0.1 & 0.1 & 0.4 & 700 \\ \hline
\multirow{4}{*}{Weather} & 96 & db2 & 2 & {[}0.2, 1.0, 0.4{]} & 0.5 & db3 & 1 & {[}0.1, 0.5{]} & 1.0 & 0.1 & 0.9 & * & * \\
 & 192 & db2 & 1 & {[}0.1, 0.7{]} & 0.5 & db3 & 1 & {[}0.2, 0.7{]} & 1.0 & 0.1 & 0.1 & * & * \\
 & 336 & db1 & 1 & {[}1.0, 1.0{]} & 1.0 & db2 & 1 & {[}0.8, 0.6{]} & 1.0 & 0.1 & 0.1 & * & * \\
 & 720 & db2 & 1 & {[}1.0, 0.8{]} & 0.5 & db1 & 1 & {[}0.1, 0.1{]} & 1.0 & 0.9 & 0.9 & * & * \\ \hline
\multirow{4}{*}{ILI} & 24 & db25 & 1 & {[}0.4, 0.8{]} & 0.2 & db1 & 1 & {[}0.1, 0.8{]} & 0.2 & 0.2 & 0.1 & 0.7 & 200 \\
 & 36 & db25 & 1 & {[}0.6, 0.8{]} & 0.2 & db25 & 1 & {[}0.1, 1.0{]} & 0.8 & 0.1 & 0.1 & 0.3 & 300 \\
 & 48 & db2 & 1 & {[}0.2, 0.7{]} & 0.2 & db3 & 1 & {[}0.1, 1.0{]} & 1.0 & 0.1 & 0.1 & 0.9 & 300 \\
 & 60 & db25 & 1 & {[}0.2, 0.8{]} & 0.2 & db1 & 1 & {[}0.1, 0.9{]} & 0.5 & 0.1 & 0.1 & 0.7 & 1000
 
\end{tabular}}
\end{table}

\newpage

\bibliography{paper}  

\end{document}